\documentclass[runningheads]{llncs}
\usepackage{graphicx}
\usepackage{amsmath,amssymb} 
\usepackage{color}
\usepackage[width=122mm,left=12mm,paperwidth=146mm,height=193mm,top=12mm,paperheight=217mm]{geometry}
\usepackage{wrapfig}
\usepackage{url} 

\newcommand{\norm}[1]{\left\lVert#1\right\rVert}
\newcommand{\tyA}[1]{#1}
\newcommand{\yeA}[1]{#1}
\newcommand{\tyB}[1]{#1}

\begin{document}
\pagestyle{headings}
\mainmatter
\def\ECCV16SubNumber{***}  

\title{StyleHumanCLIP: Text-guided Garment Manipulation for StyleGAN-Human} 

\titlerunning{StyleHumanCLIP}

\authorrunning{Takato Yoshikawa, Yuki Endo, Yoshihiro Kanamori}

\author{Takato Yoshikawa, Yuki Endo, Yoshihiro Kanamori}

\institute{University of Tsukuba, Japan}

\maketitle

\begin{abstract}
This paper tackles text-guided control of StyleGAN for editing garments in full-body human images.
Existing StyleGAN-based methods suffer from handling the rich diversity of garments and body shapes and poses. 
We propose a framework for text-guided full-body human image synthesis via an attention-based latent code mapper, which enables more disentangled control of StyleGAN than existing mappers.
Our latent code mapper adopts an attention mechanism that adaptively manipulates individual latent codes on different StyleGAN layers under text guidance. 
In addition, we introduce feature-space masking at inference time to avoid unwanted changes caused by text inputs.
Our quantitative and qualitative evaluations reveal that our method can control generated images more faithfully to given texts than existing methods. 
\end{abstract}

\section{Introduction}
Full-body human image synthesis holds great potential for content production and has been extensively studied in the fields of computer graphics and computer vision. In particular, recent advances in deep generative models have enabled us to create high-quality full-body human images. 
StyleGAN-Human~\cite{styleganhuman22} is a StyleGAN model~\cite{stylegan19,stylegan220} unsupervisedly trained using a large number of full-body human images.
The users can instantly obtain realistic and diverse results from random latent codes, yet without intuitive control.

\begin{figure*}[t]
 \begin{flushleft}
    \begin{center}
    \includegraphics[width=0.9\textwidth]{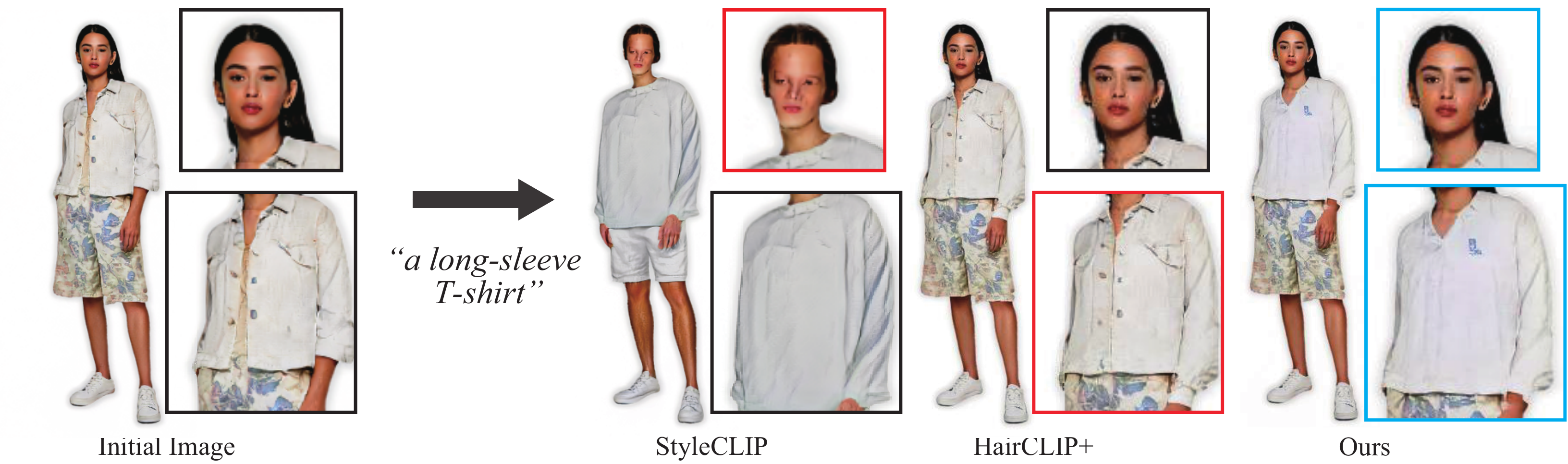}
    \end{center}
    \caption{Garment editing comparison of existing methods and ours. 
     StyleCLIP erroneously changed the facial identity and pants.
     HairCLIP+ (a HairCLIP variant trained with the same loss functions as ours) neglects the textual input due to its poor editing capability.
     Contrarily, our method successfully achieves virtual try-on of \textit{``a long-sleeve T-shirt''} while preserving the facial identity and pants.
    }
    \label{fig:teaser}
\end{flushleft}
\end{figure*}

Text-based intuitive control of image synthesis has been an active research topic~\cite{styleclip21,tedigan21,clip2stylegan22,hairclip22,diffusionclip22,stylegan-nada22,maniclip22,dalle222} since the advent of CLIP~\cite{clip21}, which learns cross-modal representations between images and texts.
StyleCLIP~\cite{styleclip21} and HairCLIP~\cite{hairclip22} can control StyleGAN images by manipulating latent codes in accordance with given texts. 
These methods succeed in editing human and animal faces but struggle to handle full-body humans due to the much richer variations in garments and body shapes and poses.
Specifically, these methods often neglect textual information on garments or deteriorate a person's identity (see Fig.~\ref{fig:teaser}). 

In this paper, we propose a StyleGAN-based framework for text-based editing of garments in full-body human images, without sacrificing the person's identity. 
Our key insight is that the existing techniques of textual StyleGAN control have a problem with the latent code mapper, which manipulates StyleGAN latent codes according to input texts.
Specifically, the modulation modules used in, e.g., HairCLIP's mapper equivalently modulate latent codes for StyleGAN layers and thus cannot identify and manipulate the text-specified latent codes.
To address this issue, we present a latent code mapper architecture based on an attention mechanism, which can capture the correspondence between a given text and each latent code more accurately.
In addition, we introduce feature-space masking at inference time to avoid unwanted changes in areas unrelated to input texts due to the latent code manipulation. 
This approach allows editing garments while preserving the person's identity. 

In summary, our contributions are that 
(i) we pioneer the issue of text-based StyleGAN image editing for full-body humans,
and (ii) we propose a framework for editing StyleGAN-Human images via an attention-based latent code mapper and feature-space masking. We demonstrate the effectiveness of our method through qualitative and quantitative comparisons with existing methods, including not only StyleGAN-based methods but also recent diffusion model-based methods. 

\section{Related Work}
\paragraph{Generative adversarial networks.}
From the advent of generative adversarial networks (GANs)~\cite{gans14}, 
various studies have explored high-quality image synthesis
by improving loss functions, learning algorithms, and network architectures~\cite{wgan17,progan18,sagan19,biggan19}. 
StyleGAN~\cite{stylegan19,stylegan220} is a milestone toward high-quality and high-resolution image synthesis. 
StyleGAN-Human~\cite{styleganhuman22} is a StyleGAN variant trained with an annotated full-body human image dataset.
However, these unconditional models lack user controllability to generate images. 

User-controllable image synthesis can be achieved via manipulation of latent codes in GANs.
For example, unsupervised approaches~\cite{DBLP:conf/nips/ChenCDHSSA16,DBLP:conf/icml/VoynovB20,DBLP:journals/corr/abs-2004-02546,DBLP:journals/corr/abs-2007-06600,DBLP:conf/iccv/HeKS21,DBLP:conf/iccv/YukselSEY21,zhu2021lowrankgan,oldfield2023panda} attempt to find interpretable directions in a latent space using, e.g., PCA and eigenvalue decomposition. 
However, finding desirable manipulation directions is not always possible.
On the other hand, supervised approaches~\cite{DBLP:conf/cvpr/ShenGTZ20,DBLP:journals/tog/AbdalZMW21,DBLP:conf/cvpr/YangCWZSH21,gansteerability,DBLP:conf/iclr/SpingarnBM21} can manipulate latent codes to edit attributes corresponding to given annotations, such as gender and age. 
However, the manipulation is limited to specific attributes,
and the annotation is costly.
We thus leverage CLIP for text-based image manipulation without additional annotations.

\paragraph{Virtual try-on.}
Recently, 2D-based virtual try-on methods~\cite{viton17,cpviton18,vtnfp19,spviton20,acgpn20,vitonhd21,hrviton22,cvton22} have been actively studied. VTON~\cite{viton17} and CP-VTON~\cite{cpviton18} are virtual try-on methods that learn the deformation and synthesis of garment images to fit target subjects. VTNFP~\cite{vtnfp19} and ACGPN~\cite{acgpn20} synthesize images better preserving body and garment features by introducing a module that extracts segmentation maps. VITON-HD~\cite{vitonhd21} and HR-VITON~\cite{hrviton22} allow virtual try-on for higher-resolution images. Although these methods require reference images of garment photographs, our method does not require reference images but uses texts as input guidance.

\begin{figure*}[t]
 \begin{flushleft}
    \begin{center}
    \includegraphics[width=0.9\textwidth]{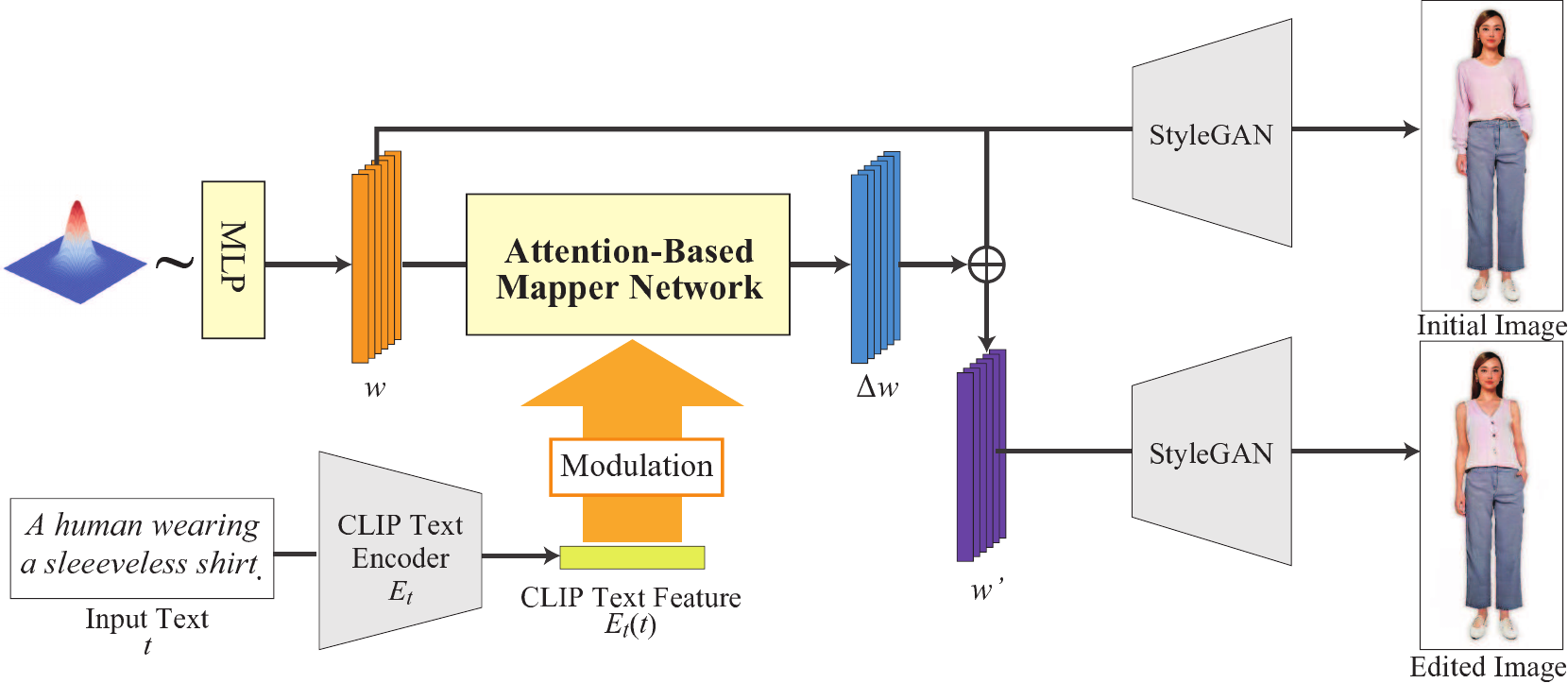}
    \end{center}
    \caption{
    Overview of the proposed framework. The mapper network translates the latent codes $w$ to the latent codes $w'$ reflecting the text input. In the training time, only the mapper network is trained, and the other networks are freezed.
    }
    \label{fig:overview}
\end{flushleft}
\end{figure*}

\paragraph{Text-guided image manipulation.}
There have been many studies on text-guided image manipulation~\cite{styleclip21,tedigan21,clip2stylegan22,hairclip22,diffusionclip22,stylegan-nada22,maniclip22,dalle222} by utilizing CLIP~\cite{clip21}. StyleCLIP~\cite{styleclip21} proposes three methods (i.e., latent optimization, latent mapper, and global directions) to edit StyleGAN images using texts. In particular, the global direction method in $\mathcal{S}$ space~\cite{stylespace21} achieves fast inference while supporting arbitrary text input. HairCLIP~\cite{hairclip22} improved the StyleCLIP latent mapper to specialize in editing hairstyles using arbitrary text input. 
However, these methods focus on editing human and animal faces and are not suitable for full-body human images due to the much richer diversity in garments and body shapes and poses.
These methods cannot appropriately reflect input texts to full-body human images and preserve the identity of face and body features. 

Diffusion model-based approaches~\cite{rombach2022high,diffusionclip22,couairon2022diffedit} have recently attracted great attention. These approaches provide high-quality editing in general domains but take more time for inference than StyleGAN-based methods. We also demonstrate that our method achieves higher-quality editing for full-body human images through comparisons with these approaches in Section~\ref{sec:exp}.

\section{Proposed Method}
Fig.~\ref{fig:overview} illustrates an overview of the proposed framework. Inspired by HairCLIP~\cite{hairclip22}, we adopt a latent code mapper trained to manipulate latent codes in the $\mathcal{W+}$ space of StyleGAN. The mapper network takes as input the latent code $w$ of the input image and the embedding $t$ of the input text and outputs the residual $\Delta w$ of the latent code. \tyA{The input $w$ is randomly sampled from Gaussian noise via the StyleGAN mapping network, and $t$ is obtained using the CLIP text encoder~\cite{radford2021learning}.}
Finally, we add $\Delta w$ to $w$ to create the edited latent code $w'$, which is fed to the pre-trained StyleGAN to obtain an edited image. 
In the following sections, we describe the architecture of our latent code mapper (Section~\ref{subsec:mapper_architecture}), training loss functions (Section~\ref{subsec:loss}), and feature-space masking in the mapper (Section~\ref{subsec:ITAN}).

\subsection{Mapper network architecture}
\label{subsec:mapper_architecture}
The mapper network used in HairCLIP~\cite{hairclip22} has several blocks consisting of a fully connected layer, modulation module, and activation function. The modulation module is used to modulate normalized latent code features via multiplication and subtraction of CLIP text features. 
HairCLIP uses three mappers (coarse, medium, and fine) to handle different semantic levels of a latent code fed to each StyleGAN layer. However, the modulation modules in each mapper equivalently modulate given latent codes. Therefore, each mapper cannot identify and manipulate only latent codes related to input texts. As a result, the HairCLIP mapper cannot reflect input texts well for full-body human images. 

\begin{figure}[t]
  \centering
  \begin{minipage}[t]{0.45\textwidth}
    \centering
    \includegraphics[keepaspectratio, scale=0.75]{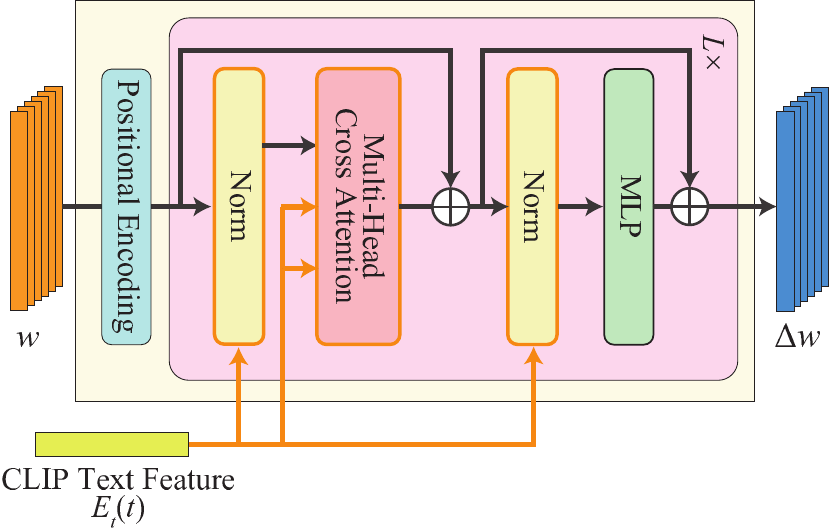}
    \caption{
    Architecture of our latent code mapper. 
    }
    \label{fig:mapper_architecture}
  \end{minipage}
  \hfill
  \begin{minipage}[t]{0.45\textwidth}
    \centering
    \includegraphics[keepaspectratio, scale=0.75]{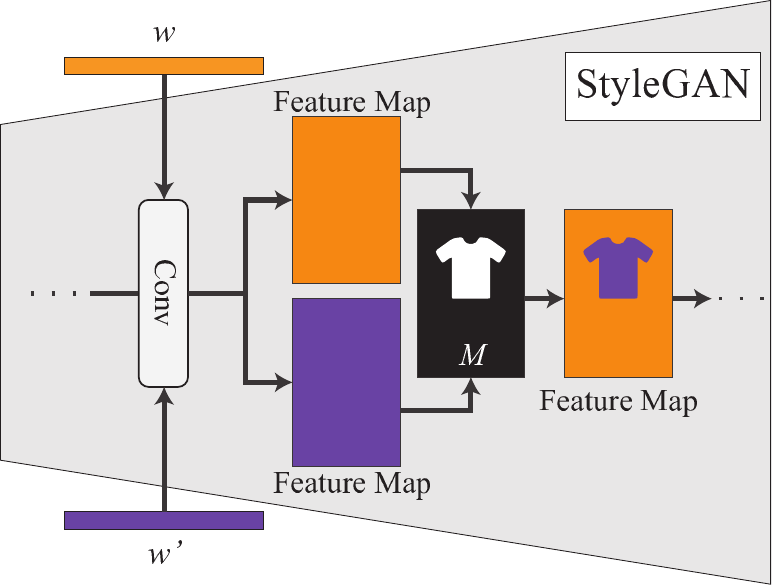}
    \caption{
    Overview of feature-space masking.
    }
    \label{fig:ITAN}
  \end{minipage}
\end{figure}

To manipulate appropriate latent codes according to text input,
we introduce a cross-attention mechanism into our latent code mapper. Fig.~\ref{fig:mapper_architecture} shows our network architecture. Our network first applies positional encoding to distinguish between latent codes fed to different StyleGAN layers. In the Norm layers, we also apply the modulation module used in HairCLIP after normalization with the mean and standard deviation of the intermediate output. In addition, following the Transformer architecture~\cite{transformer17}, we adopt the multi-head cross-attention mechanism, which can capture multiple relationships between input features. To compute the multi-head cross attention, we define the query $Q$, key $K$, and value $V$ as follows:
\begin{equation}
    Q = X_wW^Q, \hspace{1em} K = E_t(t)W^K, \hspace{1em} V = E_t(t)W^V, 
\end{equation}
where the query $Q$ is computed from the latent code feature $X_w \in \mathbb{R}^{N \times 512}$ ($N$ is the number of StyleGAN layers taking latent codes), and the key $K$ and value $V$ are computed from the CLIP feature $E_t(t) \in \mathbb{R}^{1 \times 512}$ of the input text $t$. The tensors $W^Q, W^K, W^V \in \mathbb{R}^{512 \times 512}$ are the weights to be multiplied with each input. Using the query $Q_i$, key $K_i$, and value $V_i$ for a head $i$, the multi-head cross attention is defined as:
\begin{gather}
    MultiH\!ead(Q, K, V) = [So\!f\!tmax(\frac{Q_iK_i^T}{\sqrt{d}})V_i]_{i=1:h}W^o, 
\end{gather}
where $d=512/h$ ($h$ is the number of heads), and $W^o \in \mathbb{R}^{512 \times 512}$ is the weight to be multiplied with the concatenated attentions of the multiple heads. 
\tyB{Note that, unlike the typical multi-head cross attention, our method applies the softmax function along the column direction to ensure that the weights for all latent code features sum to 1.}
We repeat the block consisting of the Norm, cross attention, and MLP layers $L$ times, as illustrated in Fig.~\ref{fig:mapper_architecture}.

\subsection{Loss functions}
\label{subsec:loss}
In the mapper network, we aim to acquire latent codes capable of generating images reflecting the input text while preserving unrelated areas. We first adopt the CLIP loss following the approach of StyleCLIP~\cite{styleclip21}.
\begin{equation}
    \mathcal{L}_{clip} = 1 - \cos(E_i(G(w')), E_t(t)),
\end{equation}
where $\cos(\cdot, \cdot)$ denotes the cosine similarity, $E_i$ and $E_t$ are the image and text encoders of CLIP, respectively, and $G(w')$ is the image generated from the edited latent code $w'$.
In addition, we introduce the directional CLIP loss presented in StyleGAN-NADA~\cite{stylegan-nada22}.
\begin{gather}
    \mathcal{L}_{direct} = 1 - \frac{\Delta T \cdot \Delta I}{\norm{\Delta T} \norm{\Delta I}},
\end{gather}
where $\Delta T = E_t(t) - E_t(t_{source})$ and $\Delta I = E_i(G(w')) - E_i(G(w))$.
One of the purposes of the directional CLIP loss in StyleGAN-NADA is to finetune the StyleGAN to avoid mode collapse caused by the CLIP loss.
Meanwhile, our method does not finetune StyleGAN, but the directional CLIP loss encourages the mapper not to train many-to-one mapping between latent codes and has an important role in generating diverse results. 
Besides, we define the background loss to ensure that areas unrelated to texts do not change: 
\begin{equation}
    \mathcal{L}_{bg} = \left\lVert (\bar{P_{t}}(G(w)) \cap \bar{P_{t}}(G(w'))) * (G(w) - G(w'))\right\rVert_2, 
\end{equation}
where $\bar{P}_{t}(G(w))$ is the binary mask representing the outside of target garment areas extracted using the off-the-shelf human parsing model~\cite{li2020self}, and $*$ denotes element-wise multiplication.
Finally, to maintain the quality of the generated image, we introduce the L2 regularization for the residual of latent codes $\Delta w$.
\begin{equation}
    \mathcal{L}_{norm} = \left\lVert \Delta w \right\rVert_2.
\end{equation}
The final loss $\mathcal{L}_{final}$ is defined as: 
\begin{gather}
    \mathcal{L}_{f\!inal} \! = \! \lambda_{c}\,\mathcal{L}_{clip} \! + \! 
    \lambda_{d}\,\mathcal{L}_{direct}
    \! + \! \lambda_{b}\,\mathcal{L}_{bg} \! + \! \lambda_{n}\,\mathcal{L}_{norm},
\end{gather}
where $\lambda_{c}, \lambda_{d}, \lambda_{b}$, and $\lambda_{n}$ are the weights for corresponding loss functions.

\subsection{Feature-space masking}
\label{subsec:ITAN}
We restrict editable areas using feature-space masking, inspired by the approach by Jakoel et al.~\cite{jakoel2022gans}. However, unlike their user-specified static masking, we have to handle masks whose shapes change dynamically according to input texts. Furthermore, there is a chicken-and-egg problem; we require a mask to generate an output image, whereas we require the output image to determine the mask shape. We solve this problem as follows. First, we generate images $G(w)$ and $G(w')$ without masking using the input latent code $w$ and the edited latent code $w'$. Second, we apply the human parsing network~\cite{li2020self} to obtain binary masks $P_{e}(G(w))$ and $P_{e}(G(w'))$ of the target garment. Finally, we merge both masks because, in case that the edited garment is smaller than the original, the original garment appears in the final image:
\begin{equation}
    M = P_{t}(G(w)) \cup P_{t}(G(w')). 
\end{equation}
Using this mask $M$, we modify a part of the StyleGAN's convolution layers and combine two feature maps created from latent codes $w$ and $w'$ during inference, as shown in Fig.~\ref{fig:ITAN}. 
By merging an input image and an edited result in the feature space, we can obtain more natural results than pixel-space masking, as discussed in Section~\ref{sec:maskeval}.

\section{Experiments}
\label{sec:exp}
\paragraph{Implementation details.}
We implemented our method using Python and PyTorch, and ran our program on NVIDIA Quadro RTX 6000.
It took about 0.3 seconds to obtain an edited image. 
The dataset contains 30,000 images randomly synthesized by StyleGAN-Human. We used 28,000 images for training and 2,000 for testing. 
For the text input, we prepared 10 text descriptions of upper-body garment shapes, 16 text descriptions of lower-body garment shapes, and 15 text descriptions of garment textures. The mapper networks were trained separately for the upper and lower bodies. Following StyleCLIP, we divided the latent codes into three groups (coarse, medium, and fine) and prepared a mapper network for each group. 
Appendix~\ref{sec:exp_details} provides more details about the training configurations.

\paragraph{Compared methods.}
We compared our method with existing StyleGAN-based methods and diffusion model-based methods. For the StyleGAN-based methods, we used StyleCLIP~\cite{styleclip21} and HairCLIP~\cite{hairclip22} combined with StyleGAN-Human~\cite{styleganhuman22}.
For StyleCLIP, we used the global direction method in $\mathcal{S}$ space~\cite{stylespace21} among the three proposed methods because it is fast and can handle arbitrary texts. 
To adapt HairCLIP to full-body human images, we changed the original loss functions designed for editing hairstyles to the same loss functions as our method. We denote this modified method as HairCLIP+. 
For diffusion model-based methods, we used Stable Diffusion-based inpainting (SD inpainting)~\cite{rombach2022high} and DiffEdit~\cite{couairon2022diffedit}. Because SD inpainting requires masks of inpainted regions, we created them using the off-the-shelf human parsing model~\cite{li2020self}. Meanwhile, DiffEdit can automatically estimate mask regions related to text inputs and edit those regions. Details on the implementation of each method are provided in Appendix~\ref{subsec:implementation_existing_methods}.

\paragraph{\yeA{Evaluation metrics.}}
As the objective evaluation metrics for quantitative comparison, we used CLIP Acc and BG LPIPS.
CLIP Acc evaluates whether edited images reflect the semantics of input texts. Inspired by the work by Parmar et al.~\cite{DBLP:journals/corr/abs-2302-03027}, we define CLIP Acc as the percentage of instances (i.e., test images) where the target text has a higher CLIP similarity~\cite{clip21} to the edited image than the input image. BG LPIPS evaluates the preservation degree of background regions outside target garment areas. We calculated LPIPS~\cite{zhang2018perceptual} between masked areas of the input and edited images. The masks are extracted using the off-the-shelf human parsing model~\cite{li2020self}. We computed CLIP Acc and BG LPIPS for 2,000 test images, which were edited using text inputs randomly selected from the prepared text descriptions. 

\subsection{\yeA{Effectiveness of our latent code mapper}}
\label{sec:quant_cmp}
\begin{figure*}[t]
 \begin{flushleft}
    \begin{center}
    \includegraphics[width=\linewidth]{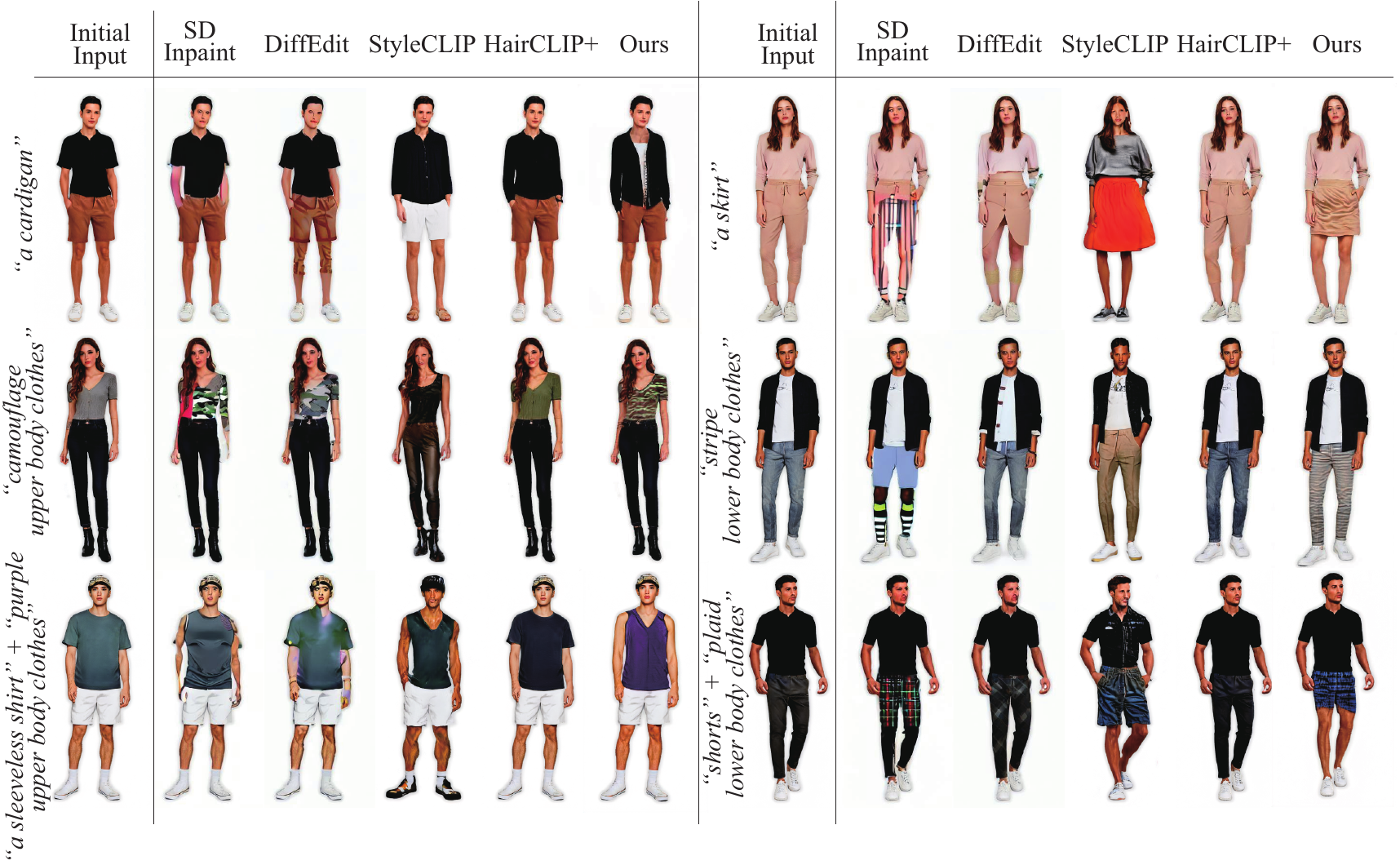}
    \end{center}
    \caption{Qualitative comparison with the existing methods~\cite{rombach2022high,couairon2022diffedit,styleclip21,hairclip22}.}
    \label{fig:qualitative}
\end{flushleft}
\end{figure*}

We first evaluate \yeA{the effectiveness of our latent code mapper} without our feature-space masking. As shown in Table~\ref{tab:quantitative_mapper}, StyleCLIP has the best score in CLIP Acc but the significantly worst score in BG LPIPS. The qualitative results in Fig.~\ref{fig:qualitative} also show that StyleCLIP changed the facial identity and garments unrelated to the text input. 
Although we can apply our feature-space masking to StyleCLIP, CLIP Acc drops significantly, as shown in Table~\ref{tab:existing_with_masking}. 
These results suggest that StyleCLIP simply increases CLIP scores by editing not only target regions but also entire regions of images, including unrelated backgrounds. The subjective user study described \tyA{in Section~\ref{subsec:user_study}} also supports this assumption. In contrast, our method has overall good scores in both metrics, which means that the edited results faithfully follow the text input while preserving unrelated areas. Finally, HairCLIP+ has the worst score in CLIP Acc, although it used the same loss functions as ours. In other words, our mapper more effectively learned text-based latent code transformation than the HairCLIP mapper in the domain of full-body human images.

\begin{table}[t]
    \centering
    \caption{
    Quantitative comparison with the existing methods. The bold and underlined values show the best and second best scores. 
    }
    \begin{tabular}{l@{\hspace{0mm}}c@{\hspace{1.5mm}}c@{\hspace{1.5mm}}c@{\hspace{1.5mm}}c} \hline
     Method & CLIP Acc $\uparrow$ & BG LPIPS $\downarrow$ \\ \hline
    StyleCLIP~\cite{styleclip21} & \textbf{98.0}$\%$ & 0.204  \\ 
    HairCLIP+~\cite{hairclip22} & 80.5$\%$ & \textbf{0.028} \\ 
    Ours w/o masking & \tyB{\underline{97.9}}$\%$  & \tyB{\underline{0.075}} \\ \hline 
    \end{tabular}
    \vspace{-2.5mm}
    \label{tab:quantitative_mapper}
\end{table}

\begin{table}[t]
    \centering
    \caption{
    Quantitative comparison with the existing methods with our feature-space masking. 
    }
    \small
    \vspace{1.5mm}
    \tyA{
    \begin{tabular}{l@{\hspace{0mm}}c@{\hspace{1.5mm}}c@{\hspace{1.5mm}}c@{\hspace{1.5mm}}c} \hline
     Method & CLIP Acc $\uparrow$ & BG LPIPS $\downarrow$ \\ \hline
    StyleCLIP~\cite{styleclip21} w/ masking & \underline{77.6}$\%$& 0.027 \\ 
    HairCLIP+~\cite{hairclip22} w/ masking & 61.1$\%$ & \textbf{0.004} \\ 
    Ours w/ masking & \textbf{\tyB{82.2}}$\%$  & \underline{0.016}}\\\hline
    \end{tabular}
    \label{tab:existing_with_masking}
\end{table}

\begin{table}[t]
    \centering
    \caption{
    User study results. Users were asked to rate alignment to text and realism of images generated by each method. }
    \tyB{
    \begin{tabular}
    {l@{\hspace{0mm}}c@{\hspace{1.5mm}}c@{\hspace{1.5mm}}c@{\hspace{1.5mm}}c} \hline
    Method & Text alignment $\uparrow$ & Realism $\uparrow$ \\ \hline
    SD Inpainting~\cite{rombach2022high} & 2.42 & 2.24  \\ 
    DiffEdit~\cite{couairon2022diffedit} & 2.10 & 2.42   \\ 
    StyleCLIP~\cite{styleclip21} & \underline{2.75} & 2.84 \\ 
    HairCLIP+~\cite{hairclip22} & 2.50 & \textbf{4.29}  \\ 
    Ours & \textbf{3.50} & \underline{4.06}  \\\hline
    \end{tabular}
    }
    \label{tab:user_study}
\end{table}

\begin{figure}[t]
  \centering
  \begin{minipage}[t]{0.48\textwidth}
    \centering
    \includegraphics[keepaspectratio, scale=0.75]{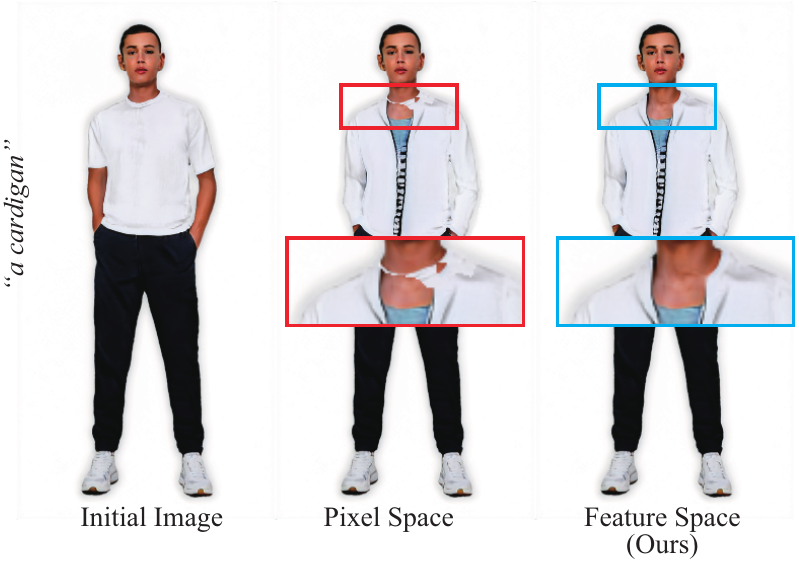}
    \caption{Qualitative comparison of pixel-space masking  and feature-space masking. }
    \label{fig:mask_method}
  \end{minipage}
    \hfill
  \begin{minipage}[t]{0.49\textwidth}
    \centering
    \includegraphics[keepaspectratio, scale=0.75]{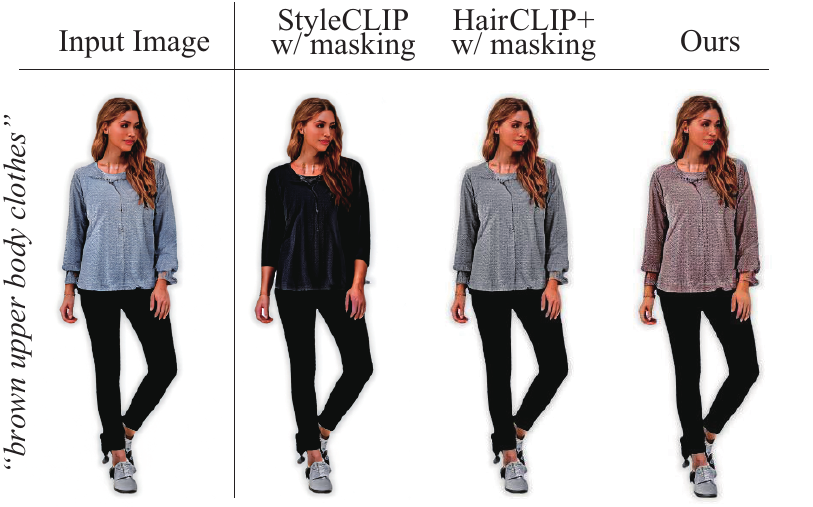}
    \caption{
    Qualitative comparison with the existing methods with feature-space masking.}
    \label{fig:existing_with_masking}
  \end{minipage}
\end{figure}

\subsection{Effectiveness of feature-space masking}
\label{sec:maskeval}
We evaluated the effectiveness of our feature-space masking. First, we compared our method with pixel-space masking, which merges target areas of edited images and the other regions of the input images in the pixel space. As shown in Fig.~\ref{fig:mask_method}, pixel-space masking yields unnatural results containing artifacts around the boundaries of garments. In contrast, feature-space masking obtains plausible results without such artifacts. 

Next, we applied feature-space masking to StyleCLIP and HairCLIP+. Fig.~\ref{fig:existing_with_masking} shows the qualitative comparison. As also mentioned in Section~\ref{sec:quant_cmp}, feature-space masking enables the existing methods to preserve areas unrelated to the specified text description, but the text input is not reflected in the outputs sufficiently. In addition, the quantitative comparisons
in Tables~\ref{tab:quantitative_mapper} and \ref{tab:existing_with_masking}
show that feature-space masking drops CLIP Acc for StyleCLIP. In contrast, thanks to our latent code mapper, our method shows the best CLIP Acc even when feature-space masking is applied. 

\subsection{\yeA{Comparison between our method and existing methods}}
Fig.~\ref{fig:qualitative} shows the qualitative comparison \yeA{between our method with feature-space masking and the existing methods}. Some results of SD Inpainting and DiffEdit effectively reflect the input text information but contain artifacts and lose fine details of faces and hands. The results of StyleCLIP in the first row show that the garment textures change together with the garment shape, even though the input text is specified to edit the shape only. In addition, the results from the second row show that StyleCLIP struggles to edit the garment textures according to the input texts. HairCLIP+ often outputs results that hardly follow the input texts. In this case, the latent code mapper of HairCLIP for face images cannot be adapted to full-body human images well. In contrast, our method correctly reflects the text semantics in the output images while preserving the unrelated areas. Please refer to Appendix~\ref{sec:additional_qualitative} for more results. 

\paragraph{User study}
\label{subsec:user_study}
We conducted a subjective user study to validate the effectiveness of our method. We asked 13 participants to evaluate 20 random sets of images edited using our method and the compared methods. The participants scored the edited images on a 5-point scale in terms of text alignment and realism. Table~\ref{tab:user_study} shows the average scores for each method. Our method obtains the best score for text alignment and 
is on par with
HairCLIP+ for realism. 
For more details on the user study, please refer to Appendix~\ref{sec:user_study_details}.

\begin{figure}[t]
    \begin{minipage}[c]{0.45\textwidth}
    \centering
    \includegraphics[keepaspectratio, scale=0.75]{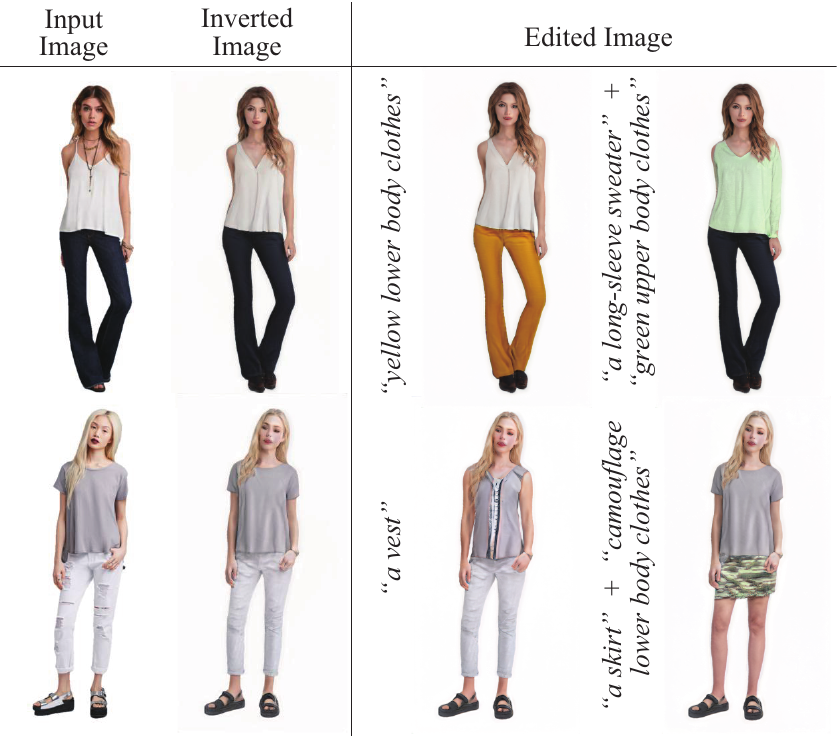}
    \caption{Application to real images. 
    }
    \label{fig:application}
    \end{minipage}
    \hfill
    \begin{minipage}[c]{0.45\textwidth}
    \centering
    \includegraphics[keepaspectratio, scale=0.75]{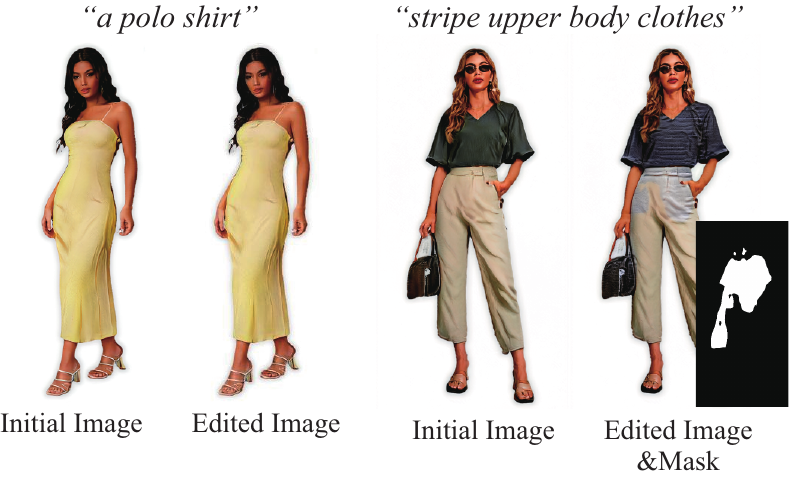}
    \caption{Failure cases. 
    Our method cannot handle full-body garments like a dress (left).
    In addition, inaccurate masks estimated by the human parsing model change unintended areas (right). }
    \label{fig:limitation}
    \end{minipage}    
\end{figure}

\subsection{Application}
As shown in Fig.~\ref{fig:application}, our method can also edit real images using GAN inversion. We used e4e~\cite{tov2021designing} to invert real images to latent codes and fed them to our mapper network. Although the inverted images lose the details of the faces and shoes, this problem arises from GAN inversion and can be alleviated by improving the inversion method.

\section{Conclusions}
To the best of our knowledge, this paper is the first attempt to control StyleGAN-Human using text input. To this end, we proposed a mapper network based on an attention mechanism that can manipulate appropriate latent codes according to text input. In addition, we introduced feature-space masking at inference time to improve the performance of identity preservation outside target editing areas. Qualitative and quantitative evaluations demonstrate that our method outperforms existing methods in terms of text alignment, realism, and identity preservation. 

\paragraph{Limitations and future work.}
Currently, our mapper networks are trained separately for the upper and lower bodies. The user needs to select the mapper networks depending on the target texts. 
In addition, we cannot handle full-body garments like a dress (see the left side of Fig.~\ref{fig:limitation}).
In the future, we want to develop a method to automatically determine which body parts should be edited according to text inputs. In addition, as shown in the right side of Fig.~\ref{fig:limitation}, our method sometimes changes unintended areas depending on
the mask $M$'s accuracy.
This problem could be improved using more accurate human parsing models.

\bibliographystyle{splncs}
\bibliography{main}

\appendix

\section{Experimental Details}
\label{sec:exp_details}
\subsection{Hyperparameter}
Our method used the pre-trained StyleGAN-Human~\cite{styleganhuman22} model, 
which has the structure of StyleGAN2~\cite{stylegan220} with a modification to output 256$\times$512 images.
We used a truncation value of $\psi=0.7$ to generate images for training and testing. The StyleGAN-Human model consists of a total of 16 layers, which are divided into three stages (i.e., course, middle, fine) with 4, 4, and 8 layers, respectively.

For our mapper network (see Section~\ref{subsec:mapper_architecture}), we set the internal block repetition count $L$ (see also Figure~\ref{fig:mapper_architecture}) to 6 and the number of heads $h$ of the multi-head cross attention~\cite{transformer17} to 4. The loss weights $\lambda_{c}, \lambda_{d}, \lambda_{b}$, and $\lambda_{n}$ were set to 1.0, 2.0, 5.0, and 1.0, respectively. We employed the Ranger~\cite{Ranger} optimizer with a learning rate of 0.0005 and (${\beta_1,\beta_2})=({0.95,0.9})$. 

To apply our method to real-world images, we utilized the e4e\footnote{\url{https://github.com/omertov/encoder4editing}}~\cite{tov2021designing} encoder trained on the SHHQ dataset containing 256$\times$512 images collected for StyleGAN-Human. For training the e4e model, we used the official default parameters, with an only modification to set the ID loss weight zero.

\subsection{Implementation of existing methods}
\label{subsec:implementation_existing_methods}
For StyleCLIP\footnote{\url{https://github.com/orpatashnik/StyleCLIP}}~\cite{styleclip21} and HairCLIP\footnote{\url{https://github.com/wty-ustc/HairCLIP}}~\cite{hairclip22}, 
we used the official implementations with a modification to replace StyleGAN with StyleGAN-Human, and reran the preprocessing and training.
For Stable Diffusion-based inpainting (SD inpainting)~\cite{rombach2022high} and DiffEdit, we used the Stable Diffusion version 1.4. For SD inpainting\footnote{\url{https://github.com/huggingface/diffusers/blob/main/src/diffusers/pipelines/stable_diffusion/pipeline_stable_diffusion_inpaint_legacy.py}}, we used the image generation pipeline of the Diffusers library. For DiffEdit\footnote{\url{https://github.com/Xiang-cd/DiffEdit-stable-diffusion/blob/main/diffedit.ipynb}}, we used the unofficial implementation because no official implementation has been released.

\subsection{Input texts}
We synthesized input texts for training by inserting labels into text templates.
Table~\ref{tab:text} shows the list of labels. 
For input text templates, we adopted ``\textit{a human wearing \{shape label\} }'' for shape manipulation and ``\textit{a human wearing \{texture label\} upper body (lower body) clothes}'' for texture manipulation. For texture manipulation, 
we randomly picked a label from the same texture label list for both upper and lower bodies. 
The input $t_{source}$ of the directional CLIP loss is set to ``\textit{a human}''.

\begin{table}[t]
    \centering
    \caption{Label list for training.}
    \begingroup
    \begin{tabular}{|c|c|c|} \hline
     Shape of upper body clothes & Shape of lower body clothes & Texture \\ \hline\hline
    \begin{tabular}{l}
        a sleeveless shirt\\
        a long-sleeve sweater\\
        a long-sleeve T-shirt\\
        a hoodie\\
        a cardigan\\
        a dress shirt\\
        a polo shirt\\
        a denim shirt\\
        a jacket\\
        a vest\\
    \end{tabular}&
    \begin{tabular}{l}
        pants\\
        slacks\\
        dress pants\\
        jeans\\
        shorts\\
        cargo pants\\
        capri pants\\
        cropped pants\\
        chino pants\\
        leggings\\
        wide pants\\
        a jogger\\
        a skirt\\
        a miniskirt\\
        a long skirt\\
        a tight skirt\\
    \end{tabular}&
    \begin{tabular}{l}
        purple\\
        red\\
        orange\\
        yellow\\
        green\\
        blue\\
        gray\\
        brown\\
        black\\
        white\\
        pink\\
        stripes\\
        dots\\
        plaid\\
        camouflage\\
    \end{tabular}

    \\ \hline
    \end{tabular}
    \endgroup
    \label{tab:text}
\end{table}

\begin{table}[t]
    \centering
    \caption{
    Selected semantic regions for mask creation.}
    \begingroup
    \begin{tabular}{c||c|c} \hline
     & Upper body & Lower body \\ \hline\hline
    Shape &
    \begin{tabular}{c}
        Upper-clothes \\ Left-arm \\ Right-arm 
    \end{tabular} &
    \begin{tabular}{c}
        Skirt \\ Pants \\ Left-leg \\ Right-leg 
    \end{tabular} 

    \\ \hline

    Texture &
    \begin{tabular}{c}
        Upper-clothes
    \end{tabular} &
    \begin{tabular}{c}
        Skirt \\ Pants
    \end{tabular} 
    \\ \hline
    \end{tabular}
    \endgroup
    \label{tab:human_parsing_label}
\end{table}

\subsection{Creating masks using human parsing model}
In our method, we use the off-the-shelf human parsing model~\cite{li2020self} to create masks for loss calculation during training and feature-space masking during inference. The human parsing model segments a full-body human image into 18 semantic regions. We create masks by selecting specific semantic regions, which differ depending on the editing areas (i.e., upper body or lower body) and the types of editing (i.e., shape or texture). Table~\ref{tab:human_parsing_label} shows the selected semantic regions in each case.

\section{Additional Qualitative Comparison}
\label{sec:additional_qualitative}
\begin{figure}[t]
    \centering
    \includegraphics[keepaspectratio, width=0.9\linewidth]{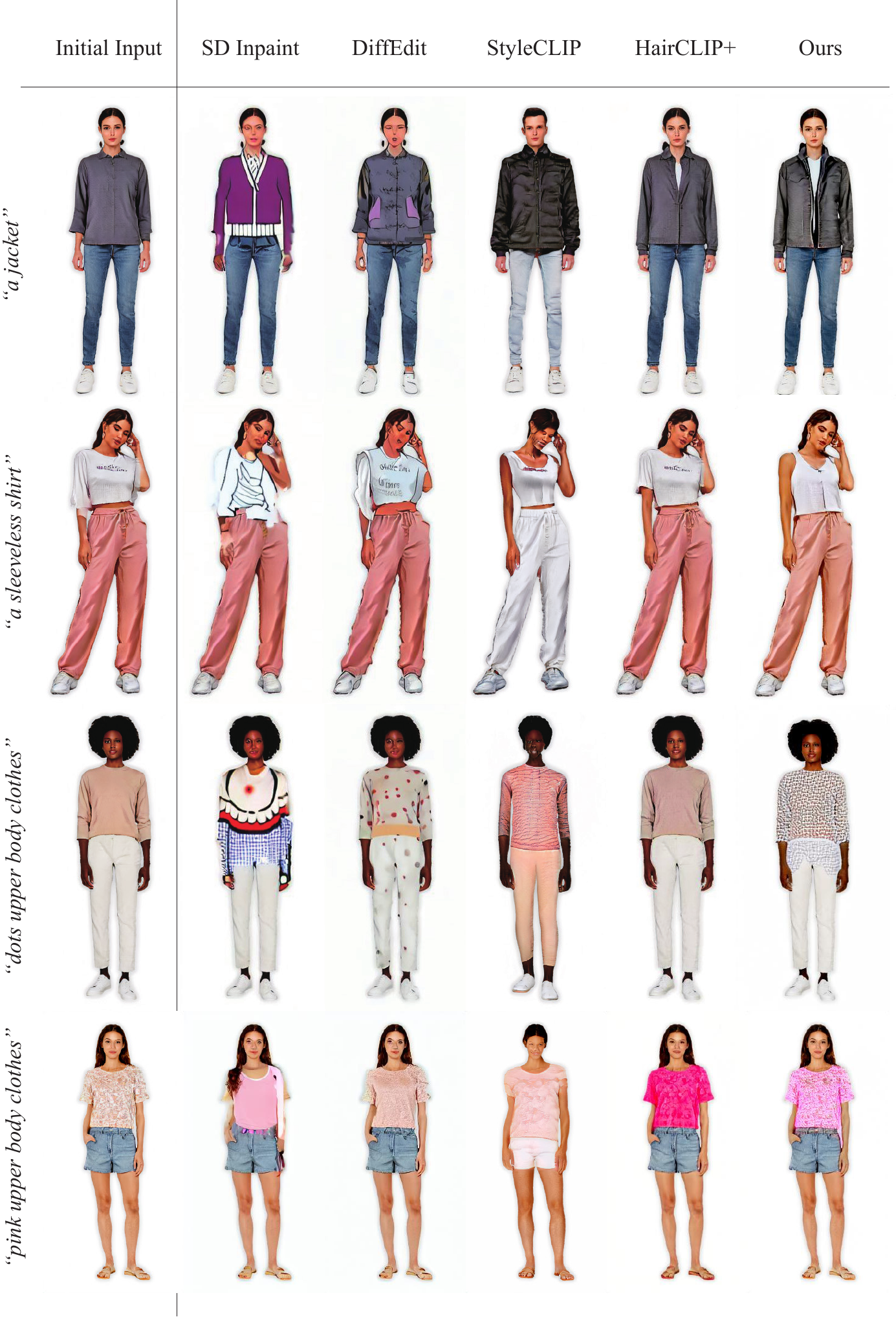}
    \caption{Additional qualitative comparison for upper body clothes manipulation.}
    \label{fig:additional_qualitative_1}
\end{figure}

\begin{figure}[t]
    \centering
    \includegraphics[keepaspectratio, width=0.9\linewidth]{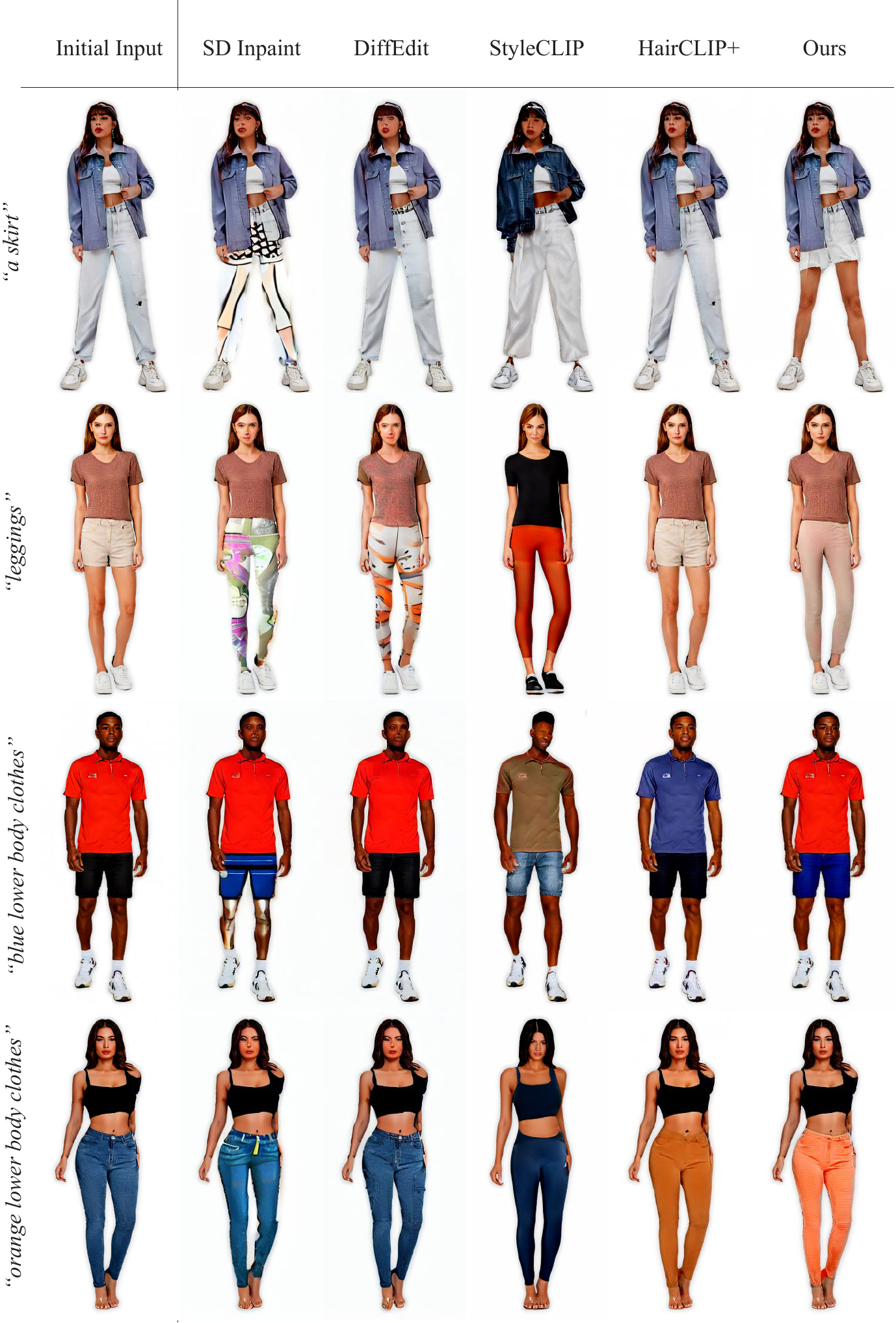}
    \caption{Additional qualitative comparison for lower body clothes manipulation.}
    \label{fig:additional_qualitative_2}
\end{figure}

\begin{figure}[t]
    \centering
    \includegraphics[keepaspectratio, width=0.9\linewidth]{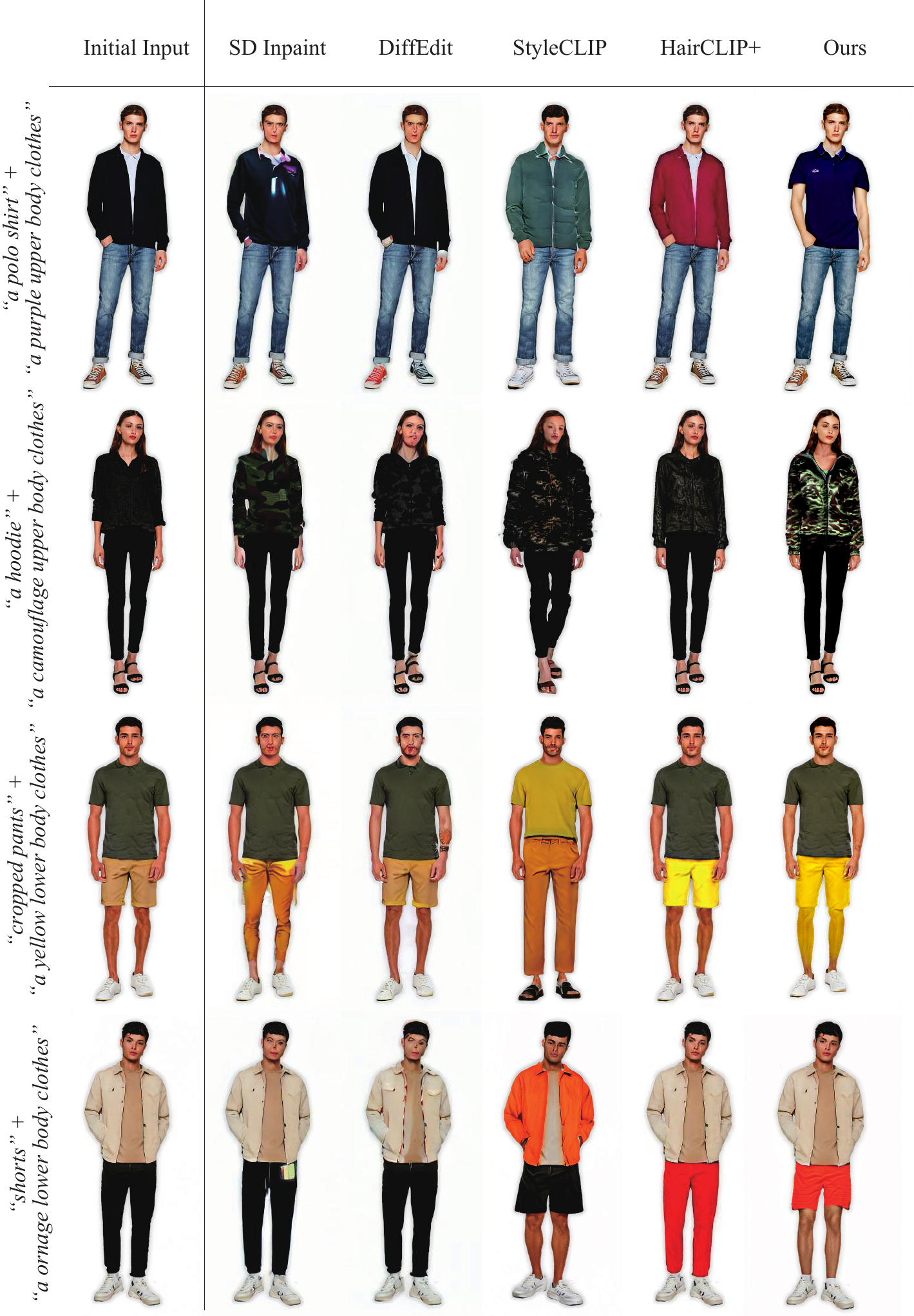}
    \caption{Additional qualitative comparison for simultaneous manipulation of shape and texture.}
    \label{fig:additional_qualitative_3}
\end{figure}

Figures~\ref{fig:additional_qualitative_1}, \ref{fig:additional_qualitative_2}, and \ref{fig:additional_qualitative_3} show the additional qualitative comparisons. Some results of SD Inpainting and DiffEdit effectively reflect the input text information but contain artifacts and lose fine details of faces and hands. The results of StyleCLIP in the first row in Fig.~\ref{fig:additional_qualitative_1} show that the garment textures change together with the garment shape, even though the input text is specified to edit the shape only. In addition, the results from the third and forth row in Figures~\ref{fig:additional_qualitative_1} and \ref{fig:additional_qualitative_2} show that StyleCLIP struggles to edit the garment textures according to the input texts. HairCLIP+ often outputs results that hardly follow the input texts. In this case, the latent code mapper of HairCLIP for face images cannot be adapted to full-body human images well. In contrast, our method correctly reflects the text semantics in the output images while preserving the unrelated areas.

\section{User Study}
\label{sec:user_study_details}
\begin{figure}[t]
    \centering
    \includegraphics[keepaspectratio, scale=0.5]{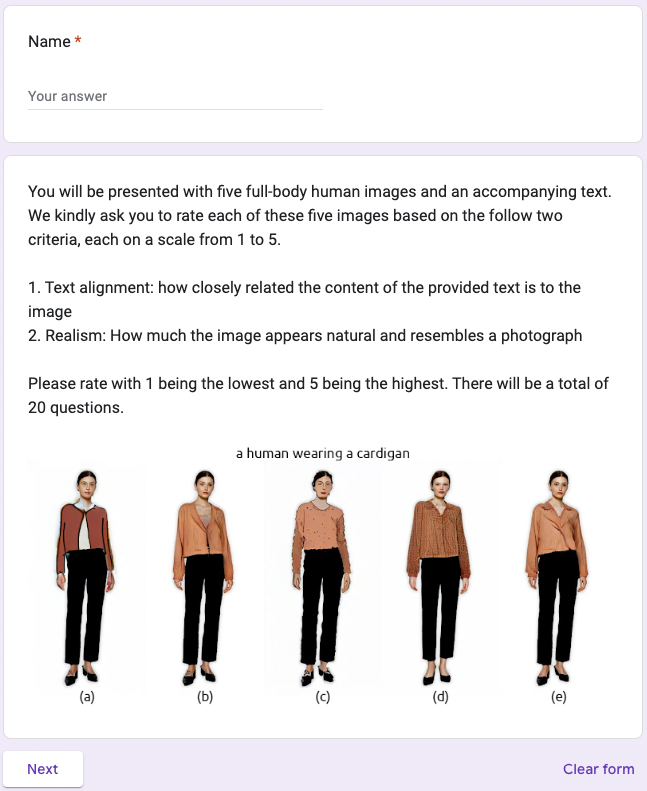}
    \caption{
    Screenshot of explanation of the user study. 
    }
    \label{fig:user_study_title}
\end{figure}

\begin{figure}[t]
    \centering
    \includegraphics[keepaspectratio, scale=0.35]{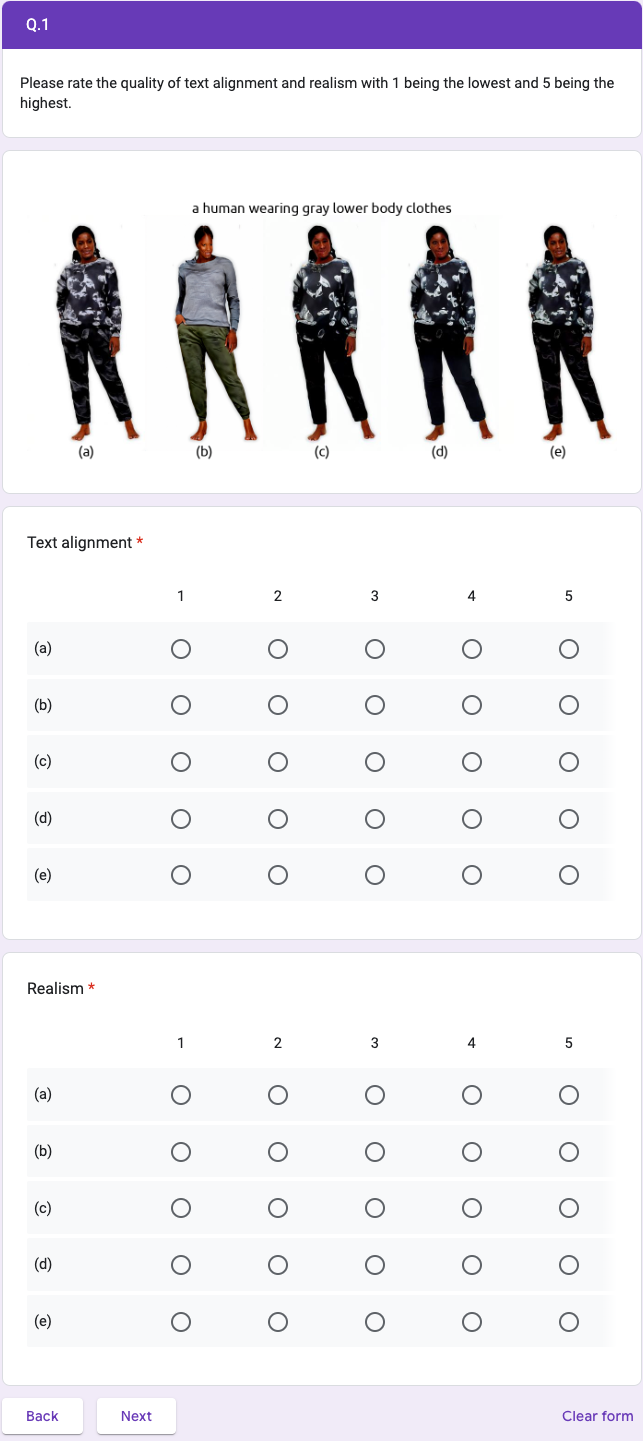}
    \caption{
    Screenshot of the first question of the user study. 
    }
    \label{fig:user_study_q1}
\end{figure}

Figures~\ref{fig:user_study_title} and \ref{fig:user_study_q1} show screenshots of the explanation and question example used in the user study.

\end{document}